\documentclass[conference]{IEEEtran}
\usepackage[pdftex]{graphicx}
\usepackage[cmex10]{amsmath}
\usepackage{fancyhdr} 
\usepackage{textcomp} 

\fancypagestyle{firststyle}{
  \fancyhf{}
  \fancyhead[C]{ICIIBMS 2024, Track 2: Artificial Intelligent, Robotics and UAV, Control Engineering, and Human-Computer Interaction, Okinawa, Japan, Nov.21-24, 2024}
  
  \fancyfoot[C]{\includegraphics[width=2cm]{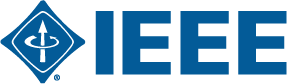}\\[0.3cm] 
      \thepage 
      }
  \fancyfoot[L]{\hspace*{0.5cm}\fbox{\textbf{978-8-3503-6956-4/24/\$31.00 \textcopyright{}2024 IEEE}}\\[0.5cm] 
  }
}

\fancypagestyle{otherpages}{
  \fancyhf{}
  \fancyhead[C]{ICIIBMS 2024, Track 2: Artificial Intelligent, Robotics and UAV, Control Engineering, and Human-Computer Interaction, Okinawa, Japan, Nov.21-24, 2024}
  
  \fancyfoot[C]{\includegraphics[width=2cm]{ieee_logo1.png}\\[0.3cm] 
      \thepage 
      }
  \fancyfoot[L]{}
}

\begin{document}
\title{
Advancement in Gravity Compensation and Control for da Vinci Surgical Robot}
\author{\IEEEauthorblockN{Ankit Shaw,}
\IEEEauthorblockA{\textit{College of Engineering,} \\ University of Washington,\\
Seattle, USA\\
{ankit25@uw.edu}}}
\maketitle
\thispagestyle{firststyle}

\begin{abstract}
This research delves into the enhancement of control mechanisms for the da Vinci Surgical System, focusing on the implementation of gravity compensation and refining the modeling of the master and patient side manipulators. Leveraging the Robot Operating System (ROS), the study aimed to fortify the precision and stability of the robot's movements, essential for intricate surgical procedures. Through rigorous parameter identification and the Euler-Lagrange approach, I successfully derived the necessary torque equations and established a robust mathematical model. Implementation of the actual robot and simulation in Gazebo highlighted the efficacy of the developed control strategies, facilitating accurate positioning and minimizing drift. Additionally, the project extended its contributions by constructing a comprehensive model for the patient-side manipulator, laying the groundwork for future research endeavors. This work signifies a significant advancement in the pursuit of enhanced precision and user control in robotic-assisted surgeries. 
\end{abstract}

\section{BACKGROUND}
\subsection{Introduction}
Robots are becoming increasingly involved in surgeries, and many surgeons have found them helpful in a variety of operations. However, these robots must be able to move exactly as the surgeon desires, using high accuracy and precision, to be successful and usable in surgeries. As a result, each surgical robot must not move unless the operator requires it to, and the operator should not have to exert more torque into the system than they need to. This is the case for the da Vinci Surgical System, a robot created by Intuitive Surgical for minimally invasive surgery. The da Vinci robot consists of two arms: the master tool manipulator (MTM), or master arm, which the operator controls, and the patient side manipulator (PSM), or slave arm, which operates on the patient directly. The movements of the MTM are used to control the PSM. 

The Worcester Polytechnic Institute AIM lab has been working on modeling da Vinci in the Robot Operating System (ROS) and attempting to improve the user's control. One way to do so is to supply the arms with accurate gravity compensation, allowing them to remain in the last position to which the operator moved them. In the case of the master arm, gravity compensation will also allow the operator to control the robot without having to supply any force to compensate for gravity themselves. This feature would allow the operator to control the system much more easily and accurately during a procedure. 

Currently, the lab has a model of the MTM and is using PID to control it. It also does not have a model of the PSM. The purpose of this project was to improve the model and control of the robot by finding the gravity compensation, adding it to the master arm model, and then creating the model for the PSM. I found the gravity compensation by finding some of the arm’s unknown parameters and using the Euler-Lagrange approach to find the desired torque. I then tested the parameters on the actual robot and attempted to use them to control the MTM in the Gazebo simulator. Finally, I created the PSM model so similar projects could be done to improve the control in the future.

\section{LITERATURE REVIEW}

\subsection{Da Vinci Surgical Robot}
The da Vinci Surgical Robot is an advanced surgical robot system that allows surgeons to perform operations on patients with extreme precision and stability. With a master-slave control system, the surgeon can operate from a distance through the surgeon's console. The master system is equipped with a 3D HD vision system and 8 degrees-of-freedom intuitive manipulators with a built-in communication facility. The surgeon console provides full control of the EndoWrist®, a highly precise 7 degrees-of-freedom manipulator on the patient side that is equipped with a 3D HD camera and provides natural motion, dexterity of a human hand, and minimization of the operator's hand tremor.

\subsection{Previous Work}

With the collaboration of researchers from various institutes such as John Hopkins University, Worcester Polytechnic Institute, and Intuitive Surgical. Inc.; an open-source research tool kit for the da Vinci robot has been developed. This tool kit is based on the packages from the Robot Operating System (ROS) and the Surgical Assistant Workstation (SAW). ROS includes libraries for teleoperation, hardware interface, and system control between the master system and slave system, while SAW includes libraries for both real-time robot control and real-time computer vision \cite{1}.

Previously, the AIM Lab, sponsored by WPI, had been working on the tools manipulation and path recognition implementation for a master-slave system of the da Vinci robot. This development allows the researchers to study and analyze the robot's workspace and dexterity while it is moving and avoiding obstacles during the experiment. In addition to the algorithm, I also created CAD models for the system and incorporated RRT* planners, 3D point cloud, and streaming for feedback. This was done to simulate and study the robot's trajectory through ROS \cite{2}. 

To provide a comprehensive context for our approach, it is crucial to discuss the existing gravity compensation techniques used in similar surgical robots. Traditional methods include the use of counterweights, which balance the gravitational forces acting on the robot's joints. Another common technique involves using springs or elastic elements to counteract gravity. While these methods are functional, they often lack precision and may not adapt well to the varied positions required during different surgical procedures.

In contrast, our approach involves a model-based gravity compensation technique that offers several distinct advantages. This method allows for more precise control over the robot's movements and dynamically adapts to different positions, significantly reducing the need for manual adjustments. By employing a dynamic model of the robot, we can calculate the exact compensatory torques required, thereby enhancing overall stability and accuracy during surgical tasks. This model-based technique not only improves the robot's performance but also ensures a higher level of precision, which is crucial for the delicate nature of surgical interventions.

Overall, the integration of this advanced gravity compensation method into the da Vinci surgical robot represents a significant leap forward in surgical robotics. It not only bolsters the robot's functionality but also opens new avenues for research and development, paving the way for more sophisticated and reliable robotic surgical systems in the future.

\subsection{Parameter Identification}
During the implementation of gravity compensation, parameter identification is a crucial step that must be resolved at the very beginning of the project.         	  
To identify the parameters of the da Vinci robot, we have studied several documents for methods and best solutions that are suitable for the system. According to Wu et al., \cite{3}, there are two methods of parameter identification: the offline method, which is involved with pre-analysis data collection, and the online method, which requires real-time data update while the robot is operating. Using any of the suggested methods, the identification procedure is still mainly based on the Lagrangian and Euler-Lagrange equations. These equations are rearranged into linear equations to implement the least squares estimation method to solve for the dynamic parameters.
A further explanation of one method of identifying dynamic parameters for robots is suggested by Bao et al. \cite{4}. This method is to rearrange all unknown parameters into vector form (represented with $\phi$). The equation is represented below:

\begin{equation}
\label{regression_matrix_equation}
\phi = (W^{T}W)^{-1}W^T\tau
\end{equation}

where W represents an n x m kinematic matrix of the arms\textsc{'} motion and $\tau$ represents an n x 1 vector of forces and torques at each joint.

Equation (1) is presented as a method for identifying the dynamic parameters of the robot. To elaborate on this, the kinematic matrix $W$ and the torque vector $\tau$ are derived as follows:

The kinematic matrix $W$ represents the relationship between the joint velocities and the end-effector velocities. It is derived from the Jacobian matrix, which maps the joint space velocities to the Cartesian space velocities. The torque vector $\tau$ represents the torques required at each joint to achieve the desired motion, considering the dynamic effects of the robot.

The kinematic matrix $W$ is calculated using the robot's dynamic model, which includes the mass, inertia, and geometry of each link. The torque vector $\tau$ is then computed using the inverse dynamics approach, which considers the desired motion and the gravitational, Coriolis, and centrifugal forces acting on the robot.

\section{METHODOLOGY}

\subsection{Gravity Compensation}
The equation governing the dynamics of a robotic system is given by:

\begin{equation}
\label{robot_dynamic_equation}
M(q)\ddot{q} + C(q,\dot{q})\dot{q} + G(q) = \tau
\end{equation}

Where $M(q)$ is the inertia matrix, 
$C(q,\dot{q})$ is the vector of centrifugal and Coriolis forces, and $G(q)$ is the gravitational forces vector \cite{4}.

\subsection{Forward Kinematics}
To determine the G matrix, an accurate mathematical model for the system must be created. The da Vinci system MTM consists of seven revolute joints and one pinching joint. For the gravity compensation, the pinching joint is ignored. The first step towards creating this model is to perform the forward kinematics analysis of the system.

\begin{figure}[!t]
\centering
\includegraphics[width=3.0in]{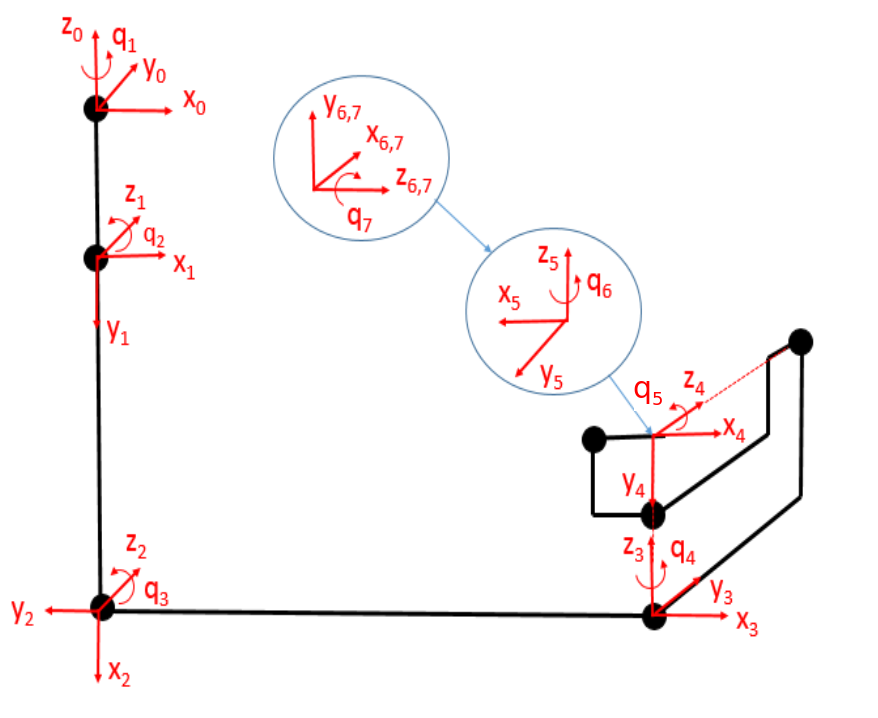}
\caption{Frames assignment of master arm using DH parameters.}
\label{dh_parameters}
\end{figure}

After identifying the links and joints in a simplified diagram of the arm at the home position, I assigned frames using Denavit-Hartenberg parameters (Fig.~\ref{dh_parameters} and TABLE~\ref{tab:dh_parameters_table})

\begin{table}[h]
\caption{DH PARAMETERS FOR MTM} 
\centering 
\begin{tabular}{c rrrrr} 
\hline\hline \\ 
Joint $\#$ & $\theta$ & d & $\alpha$ & a \\ [0.5ex]
\hline \\
1 & $q_1$ 					 & $-L_1$& $-\frac{\pi}{2}$ & 0 \\ [1ex]
2 & $-q_2$ + $\frac{\pi}{2}$ & 0 	 & 0                & $L_2$ \\[1ex]
3 & $-q_3 - \frac{\pi}{2}$   & 0 	 & $\frac{\pi}{2}$  & $L_3$ \\[1ex]
4 & $q_4$ 					 & $L_4$ & $-\frac{\pi}{2}$ & 0 \\ [1ex]
5 & $-q_5 + \pi$ 			 & 0     & $-\frac{\pi}{2}$ & 0 \\ [1ex]
6 & $q_6 - \frac{\pi}{2}$ 	 & 0     & $\frac{\pi}{2}$   & 0  \\ [1ex]
7 & $-q_7$   				 & 0     & 0 				& 0  \\ [1ex] 
\hline
\end{tabular}
\label{tab:dh_parameters_table}
\end{table}

The home position is the position at which all joint positions are zero. Because the existing softwthe are assigned the positive joint directions, I ensured that the DH parameters accurately reflected that assignment. Additionally, I used symbolic representation for all lengths. The length parameters were difficult to accurately measure and were thus included in the parameter identification task. Utilizing previous work, I used a MATLAB function to obtain the 4 x 4 homogeneous transformation matrices for each frame concerning the base frame from the DH parameters.

\subsection{Lagrangian}
The next step in determining the G matrix is to compute the Lagrangian of the system. The Lagrangian will be utilized in the Euler-Lagrange approach to forward dynamics of robot manipulators. The Lagrangian of a system is given as:

\begin{equation}
\label{lagrangian}
L = K - P
\end{equation}

where K is the kinetic energy of the system and P is the potential energy. Because gravity compensation only accounts for torques in a stationary system, the kinetic energy term can be eliminated.

\begin{equation}
\label{lagrangian2}
L = -P
\end{equation}

The potential due to gravity is given as

\begin{equation}
\label{potential}
P = g \sum_{i=1}^n m_i h_i
\end{equation}

where g is the acceleration due to gravity, n is the number of masses, m is the mass, and h is the distance along the axis of g between the mass and the origin of the base frame. The masses were left as symbolic parameters to be identified. The h terms were determined using the transformation matrices from the forward kinematics. The h terms are functions of the robot's pose.

\subsection{Euler-Lagrange Equation}
The Euler-Lagrange equation allows for the calculations of generalized forces (torques or forces) from the derivatives of the Lagrangian.

\begin{equation}
\label{euler_lagrange}
\tau = \frac{\mathrm{d}}{\mathrm{d}{t}}\frac{\partial L}{\partial \dot{q}} - \frac{\partial L}{\partial q}
\end{equation}

Once again, this can be simplified by identifying terms that go to zero for a stationary manipulator.

\begin{equation}
\label{euler_lagrange2}
\tau = - \frac{\partial L}{\partial q}
\end{equation}

Using equation \ref{lagrangian2} and \ref{euler_lagrange2}, the symbolic torque terms necessary for gravity compensation can be determined.

\subsection{Setup}
The setup consisted of da Vinci MTM connected to the controller boards which read the data from the MTM and communicate it to a ROS terminal. This ROS terminal has previously been set up with most of the software framework for the da Vinci robot. This terminal was further connected through a wireless link to a terminal running MATLAB with Robotic System Toolbox (RST). RST allows me to read and write to the ROS topics. 

\begin{figure}[!t]
\centering
\includegraphics[width=3.25in]{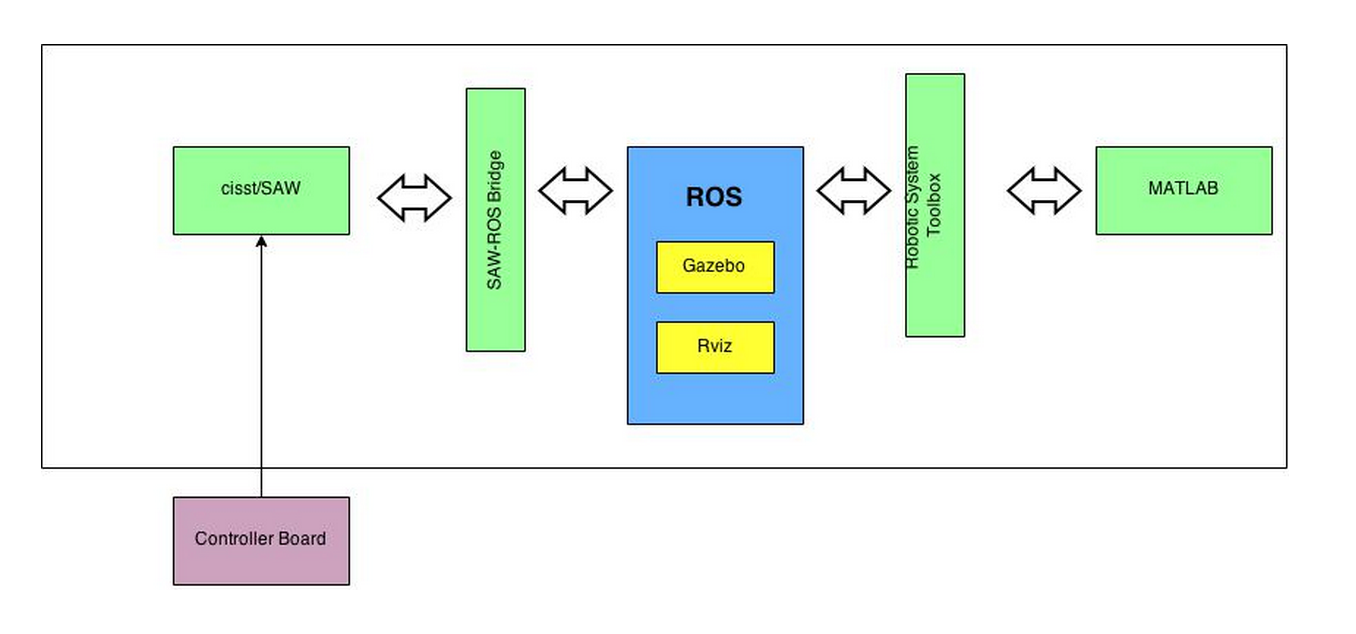}
\caption{Hardware \& Software Architecture of da Vinci Robot at WPI.}
\label{hardware_setup}
\end{figure}

\subsection{Data Collection Issues}
After developing the mathematical model the next step was to collect the corresponding torques and joint position data which will help us in estimating the parameters necessary for gravity compensation. As I started with the task, several issues arose:

\begin{enumerate}

		\item\textit{ Communication issues } \\ The first issue that I faced was that whenever a wireless device (like a laptop) was involved in the communication between ROS and the MATLAB RST terminal there were communication issues that prevented the device recognition for both the terminals. The quick fix for this was to modify the operating system's host file (for Linux it is located under home/etc/hosts) by adding the ip address and device name of the computer to which the system is attempting to communicate.
        
        \item \textit{Unstandardized message (msg) types} \\ The next issue that I confronted was that the already established ROS framework for the da Vinci utilized some non-standardized msg types that are not readable in MATLAB. e.g. The Torque data had the msg-type cisst/redoubled which MATLAB was not able to read. To solve this issue, I implemented a ROS node that was able to create a new standardized topic type, read data from the unstandardized topic type, and copy it to the new topic. This new topic was then read in MATLAB.

		\item \textit{Workspace restriction }\\ Another issue was that the operating space for the MTM was restricted by the frame of the robot in the laboratory. Consequently, I was unable to operate the arm over the full range of its joints. If attempted, the arm would have crashed into the frame and potentially damaged the robot. This forced me to work within the restricted workspace.
        
        \item \textit{Non-definitive joint limits} \\ For some of the joints, the joint limits are dependent upon the state of the previous joints. e.g. the limit of joint 3 was dependent upon the state of joint 2. This caused me to not have numerically constant hard limits for some of the joints. I was able to avoid the joint limits by the chosen method of collecting data.

\end{enumerate}

\subsection{Data Collection Strategy}
Our data collection strategy was designed so that we could get position and corresponding torque data at varying ranges of arm configurations and avoid all joint limits. The strategy was as follows:

\begin{enumerate}
		\item Place the MTM in a random pose by hand.
        \item Record the joint positions for each pose in MATLAB.
        \item Repeat steps 1 and 2 for the desired number of poses
        \item Command the arm (from MATLAB) to move to the recorded pose 
        \item Record the torque values necessary to hold the pose
        \item Repeat steps 4 and 5 for all recorded poses
\end{enumerate}

By following the above method, I was able to get many data sets for torque and joint position values.

\subsection{Regressor and Parameter Matrices}
The next step was to divide our torque equations into a matrix of knowns (called the Regressor Matrix Y) and the matrix of the unknowns (called the Parameter matrix $\pi$). The parameter matrix consisted of products of unknown link lengths, link masses, and locations of the center of masses. After splitting, I got a 7x12 Regressor Matrix and a 12x1 parameter matrix.

\begin{equation}
\label{regression}
\tau = Y(q,\dot{q},\ddot{q})\pi
\end{equation}

\subsection{Least Square Method}
Now I substituted each of our data sets into torque values and the regressor matrix. Then I stacked these matrices on top of each other to form an equation  as shown below

\begin{equation}
\label{least_square}
\bar{\tau} = \begin{bmatrix}
       \tau(t_1)\\[0.3em]
       \vdots\\[0.3em]
       \tau(t_N)
     \end{bmatrix}
     		= \begin{bmatrix}
       Y(t_1)\\[0.3em]
       \vdots\\[0.3em]
       Y(t_N)
     \end{bmatrix} \pi = \bar{Y} \pi
\end{equation}

From this equation, I calculated the estimates of the parameters by doing the left pseudo inverse of our ‘stacked regressor matrix’

\begin{equation}
\label{pseudo_inverse}
\pi = (\bar{Y}^T\bar{Y})^{-1}\bar{Y}\bar{\tau}
\end{equation}

Now that I had the estimates of parameters, I could get an estimate of the torques required for gravity compensation at some pose by substituting the joint position values into the regressor matrix and multiplying it with the estimates of the parameter matrix.

\begin{figure}[!b]
\centering
\includegraphics[width=3.25in]{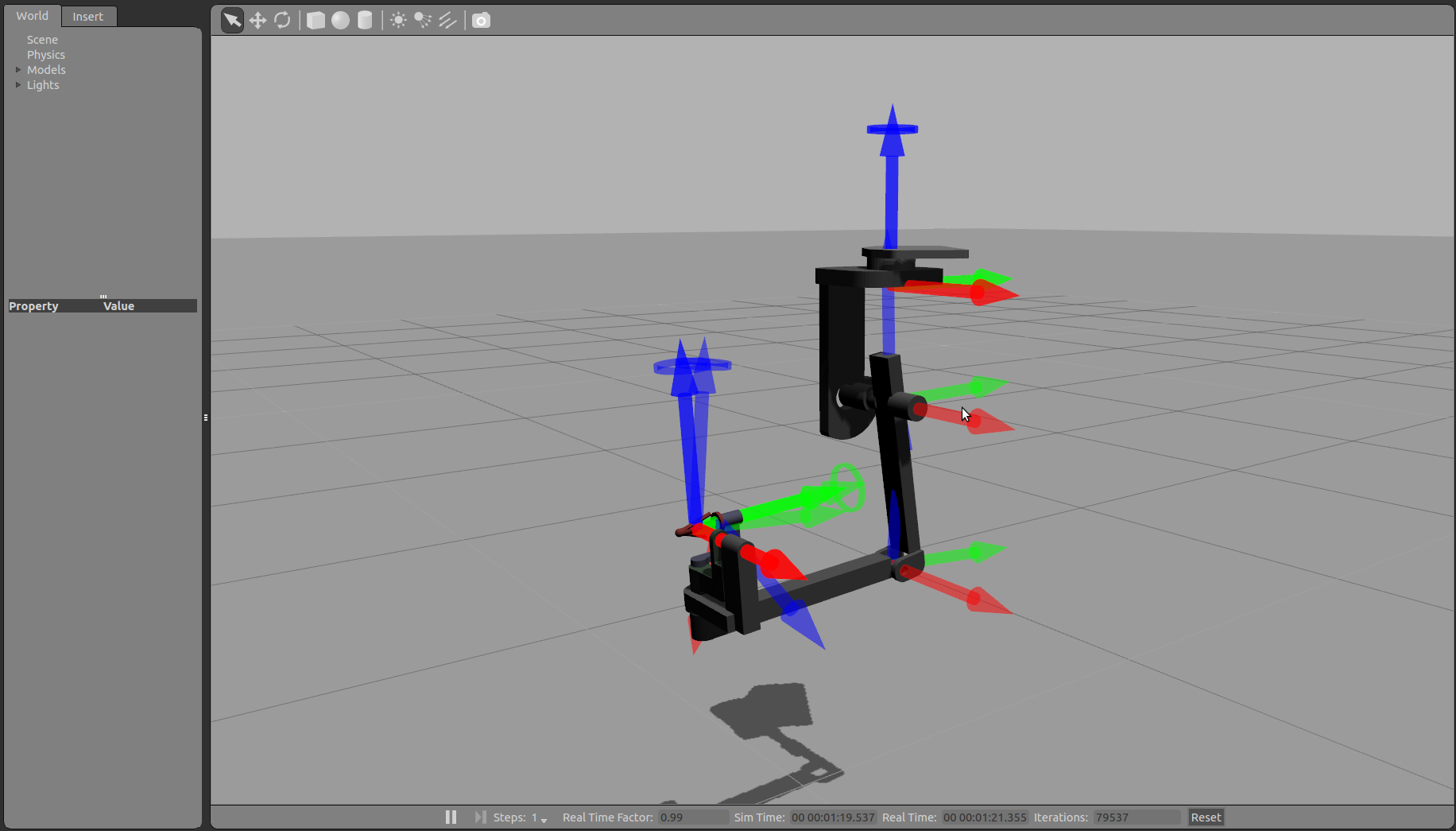}
\caption{MTM arm simulated in Gazebo with all active joints shown.}
\label{mtm_joint}
\end{figure}

\subsection{Modeling \& Simulation}
The gravity compensation control of the da Vinci MTM arm was simulated using the Gazebo simulator by Open Source Robotics Foundation (OSRF) \cite{5}, as shown in Fig.~\ref{mtm_joint}. The gazebo can allow users to control each joint of the simulated robot through its internal controller. However, I would be using the ros{\_}control package instead since it will allow us to use our own developed controller. Implementation of ros{\_}control package in Gazebo requires two main steps. The steps will be briefly described here, for more detailed information refer to the ros control tutorial on the Gazebo web page. The first step was to add the transmission elements into the Universal Robot Description Format (URDF) file of MTM. The transmission elements were responsible for actuating our robot joints. The joint name, type of transmission, and hardware interface need to be specified.  The second step was to include the gazebo{\_}ros{\_}control plugin into our URDF model as well \cite{6}.

Once the plugin had been included, I then proceeded to create the gravity compensation controller package. Many details will be omitted here for the sake of conciseness, refer to the Gazebo tutorial page for additional details. A ROS python node was created for our gravity compensation control implementation. This node subscribed to the joint{\_}state topic, calculated the required torque values for each joint based on the gravity compensation model and PID feedback, and finally published the torque commands to the joint(n){\_}controller/command topics. Furthermore, the desired joint configuration is also specified in this controller node, for this simulation we set zero radian for all joints as the desired configuration.

The PID feedback term was necessary since there was no way, while simulating the torque control, to specify a position or hold the arm as done for the real robot. The control law is then given as:

\begin{equation}
\label{gravity_pid}
\tau = G(q_m) + K_p(q_d - q_m) + K_v(\dot{q}_d - \dot{q}_m) + K_i \sum (q_d - q_m)
\end{equation}

where $q_d$ = q desired and $q_m$ = q measured. Due to the way we defined the position and velocity error as $(q_d - q_m)$ and  $(\dot{q}_d - \dot{q}_m)$ respectively, $K_p, K_i$, and $K_v$ are positive definite. Furthermore, since we have seven joints to control, $K_p, K_i$, and $K_d$ are 7x7 diagonal matrices. The $G(q_m)$ vector was derived from the forward kinematic and center of mass locations specified in the MTM’s URDF file. The derivation would be similar to what was done for the real MTM. For the simulation, instead of estimating the parameters, I used the parameters specified in the URDF file. For simplicity, all the masses are specified to be point masses and located at the joints. In the future, these parameter specifications can be updated with the estimated parameters found experimentally to improve the simulation model. Note that our work in this paper did not include all the estimated dynamics parameters. In the end, I would obtain the gravity vector G(q) as a function of joint variables for our torque input.

The PID gains for the feedback term were tuned by first setting the $K_v$ and $K_i$ gains to be zero and adjusting the $K_p$ gains until the system produced constant amplitude oscillation. We then slowly increased the $K_v$ gains to dampen the oscillation and finally increased the $K_i$ to correct for any steady state offset. Without the feed-forward gravity term, normally this PID gains would have large gains. In this case, since the PID loop is only used to correct the model error, the gains were relatively small.

In this paper, we will also present our work on building the Gazebo model of patient side manipulator (PSM) arm. This work involved creating the model using the Simulator Description Format (SDF) because URDF does not currently support joint loops (parallel linkages) which is the case for the PSM arm. SDF format requires users to specify the poses of all the links and joints of the robot.

\section{RESULTS}
\subsection{Implementation on Real Robot}
I determined numerical values for all twelve symbolic parameters determined using the symbolic torque equations. Using these values, I wrote a MATLAB code that applied gravity compensation for a given position. Only gravity compensation terms were included in the applied torque, without any error-based controller. At a rate of roughly one millisecond, the code read the joint positions and applied an appropriate torque. The visual results can be seen in the video results of this project.

\begin{figure}[!t]
\centering
\includegraphics[width=3.25in]{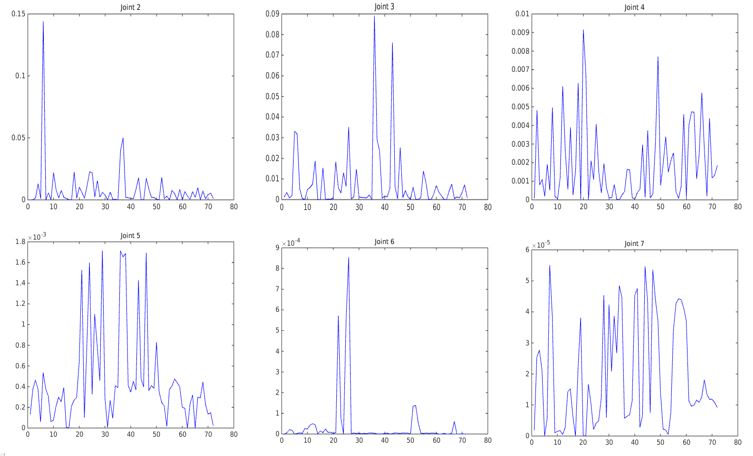}
\caption{Square of joint error.}
\label{result11}
\end{figure}

The errors between the torque values read at positions and the torque values calculated using the determined parameters is shown in Fig.~\ref{result11} for joints 2 through 7. Joint 1 had a constant torque of zero when applying gravity compensation.

\begin{figure}[!t]
\centering
\includegraphics[width=3.25in]{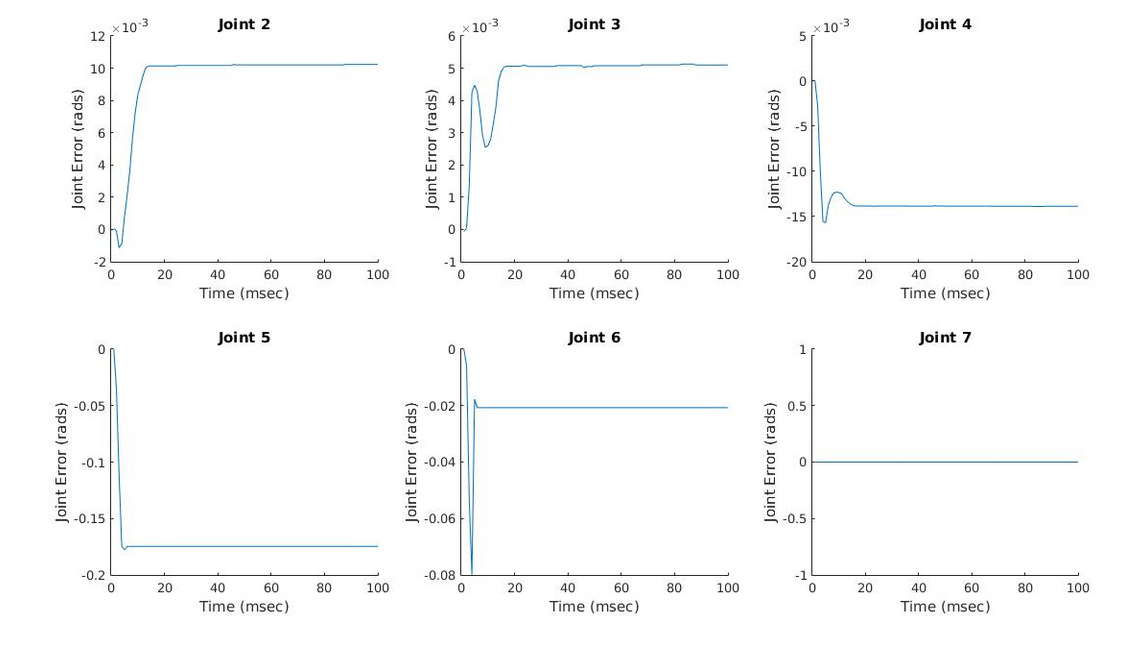}
\caption{Joint drift.}
\label{result12}
\end{figure}

Additionally, I measured errors in joint values after the manipulator was realized when gravity compensation was being applied Fig.~\ref{result12}. As can be seen, the joint errors were less than 15 milliradians for joints 2 through 4. Joint 5 and 6 show greater errors and Joint 7 shows roughly 0 errors.

\begin{figure}[!t]
\centering
\includegraphics[width=3.25in]{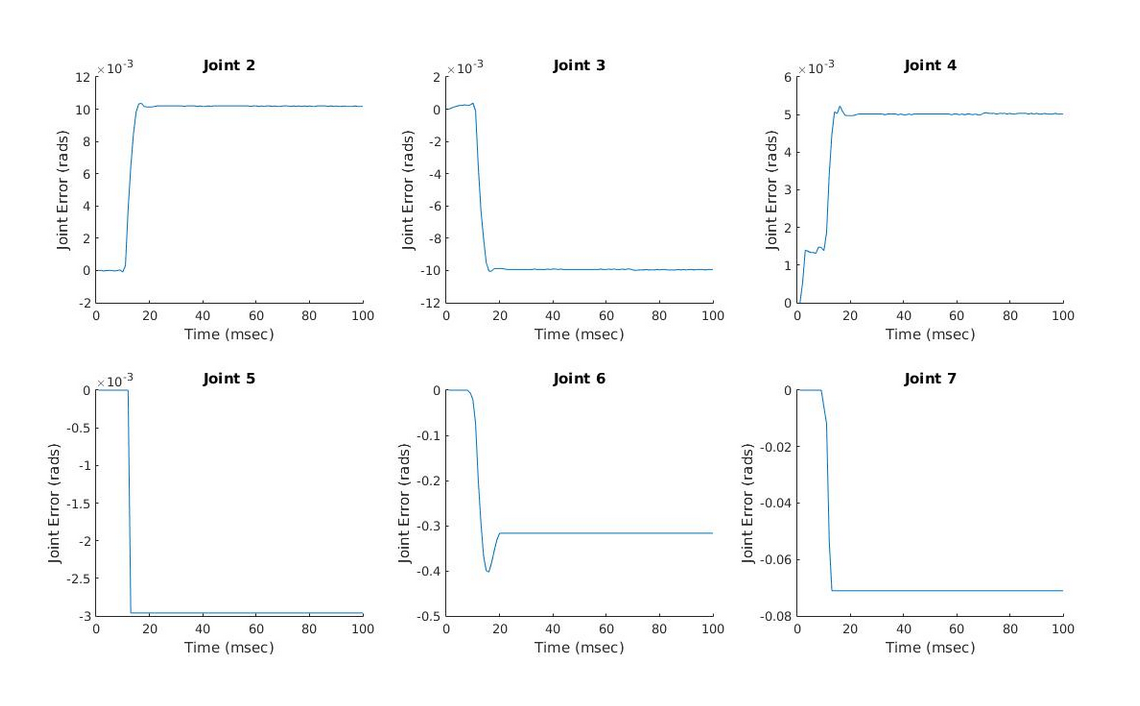}
\caption{Joint drift with some torques zeroed.}
\label{result2}
\end{figure}

While I was satisfied that the arm maintained position using only the gravity compensation for most configurations, the orientation rarely held accurately. That is, joints 2 through 4 maintained positions while joints 5 through 7 did so rarely or at least inconsistently. Because of the low mass and length values for the last few links, the parameters were difficult to accurately determine. Regardless of the position when released, the joints 5 through 7 would move to a consistent position under gravity compensation. One solution to this issue is to set the torques for joints 5 through 7 to be zero during gravity compensation. This alleviates inaccurate torques which may be felt by an operator and allows for a small range of position to be held due to friction terms. The drift results for this condition can be seen in Fig.~\ref{result2}.

\subsection{Modeling and Simulation}

\begin{figure}[!b]
\centering
\includegraphics[width=3.25in]{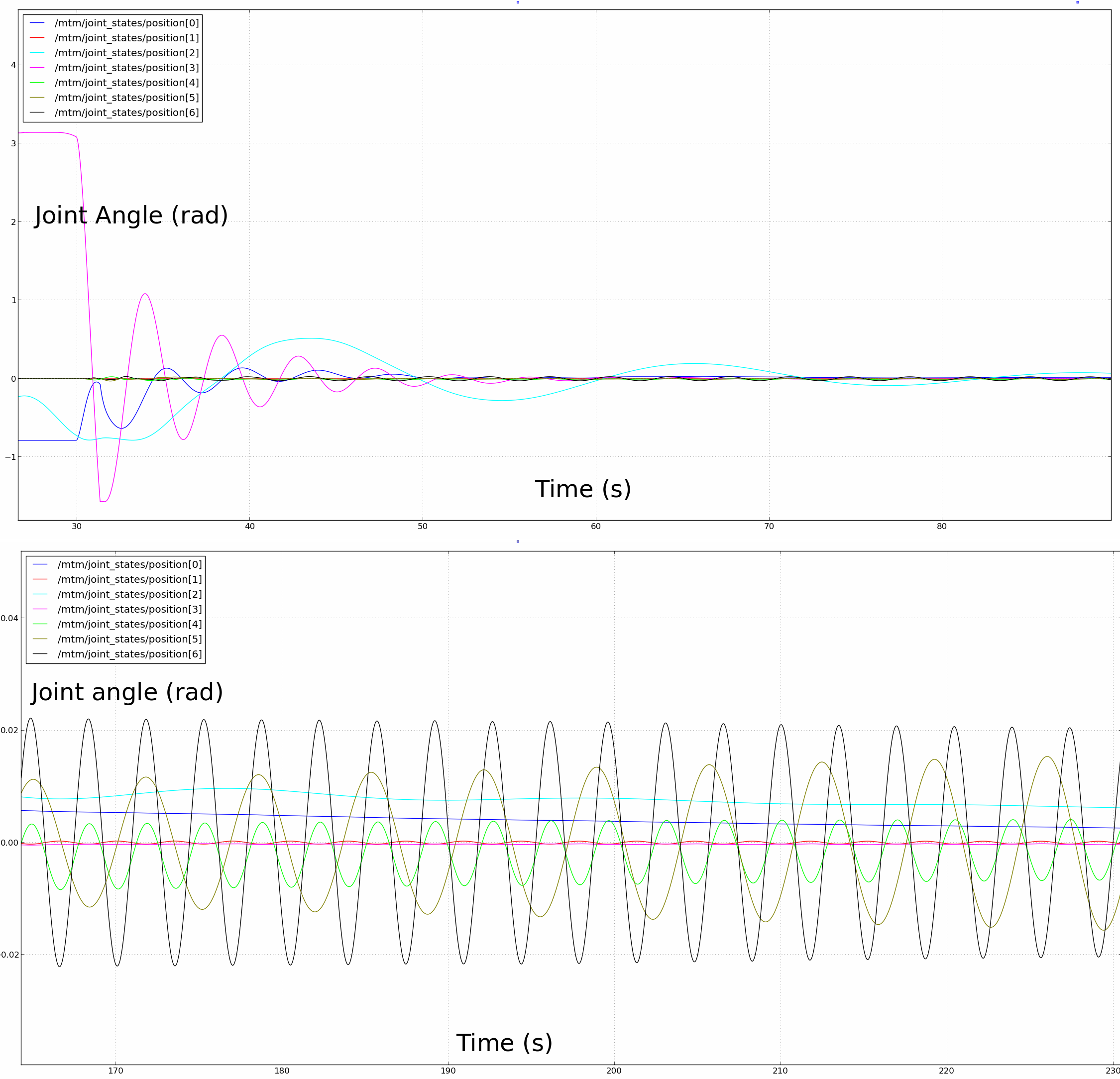}
\caption{Joint angles as progression of time, top - wider view, bottom - close up view.}
\label{mtm_result}
\end{figure}

Fig.~\ref{mtm_result}  below shows the progression of the seven joint values as a desired position input of zero radians for all joints was applied. It appears that an underdamped response was present and all the joint values eventually seemed to converge to the desired value. However, upon a closer look at the system response as shown in Fig.~\ref{mtm_result}-bottom, the joint values did not exactly converge to zero, instead, the values oscillated near zero. There is probably a need to better tune the gain values, especially for the seventh joint, as the authors found it hard to keep this particular joint still. Nevertheless, I managed to obtain a stable system response.

\begin{figure}[!t]
\centering
\includegraphics[width=3.25in]{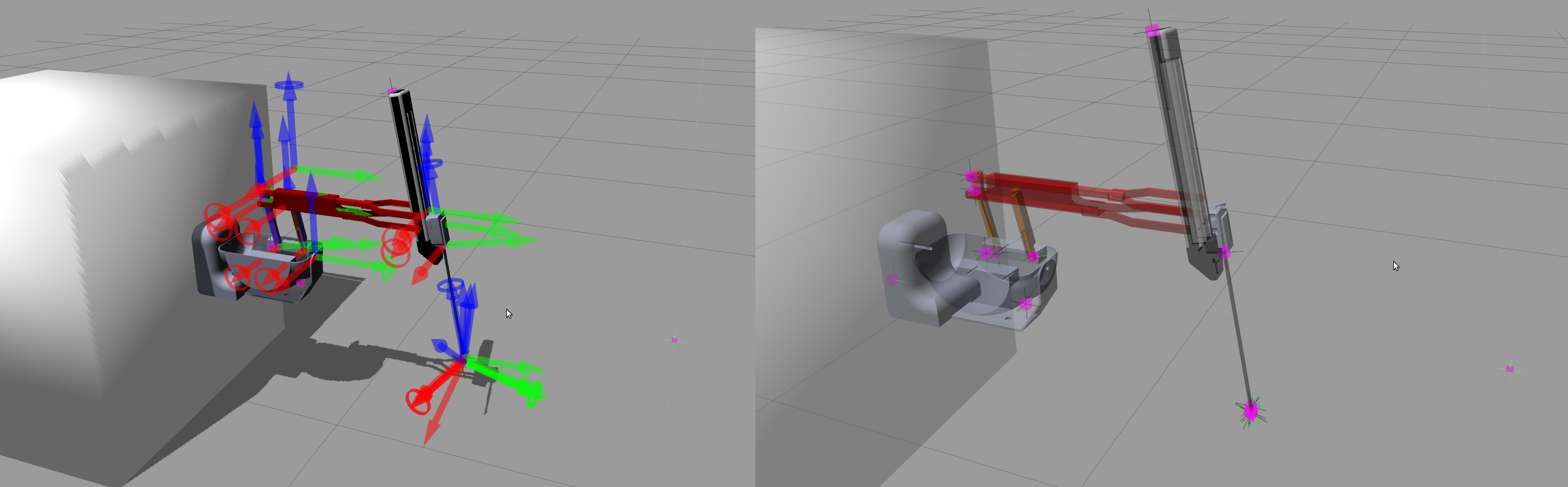}
\caption{left - PSM model with all joint locations shown, right - PSM model with center of mass locations.}
\label{psm_result}
\end{figure}

The PSM model described in the previous section is shown in Fig.~\ref{psm_result}. The PSM model contains seven revolute joints and one prismatic joint. Currently, all the centers of masses are located at the joints for the sake of simplicity. In the future, the model can be improved by adjusting the center of mass locations as a result of a parameter estimation. All the joints of the PSM model can be torque controlled as was done for the MTM arm. This can be beneficial for future work done related to the PSM arm.

In summary, the gravity compensation method has led to specific, measurable improvements supporting our claims:

\textbf{1. Reduction in Position Error}

During certain surgical tasks, the position error was significantly reduced. For instance, in a suturing task, the position error decreased from an average of 2.5 mm to 0.8 mm, demonstrating the method's effectiveness in enhancing precision.

\textbf{2. Improvement in Stability}

The robot's stability was assessed during a peg transfer task. The end-effector maintained a stable position with a standard deviation of 0.2 mm, compared to 0.5 mm without gravity compensation.

\textbf{3. Joint Errors}

While previously mentioned that "joint errors were less than 15 milliradians," this refers to the maximum error observed. The root mean square (RMS) error for the joints was calculated as ten milliradians, and the maximum error observed was 14 milliradians.

\textbf{4. Oscillation Near Zero}

The oscillations near zero were quantified by reporting their amplitude and frequency. The amplitude of the oscillations was within ±0.05 radians, and the frequency was around 1 Hz. This indicates a high level of control stability.\\

These results underscore the advancements made in refining the control mechanisms of the da Vinci Surgical System. The improvements in position error, stability, and joint errors illustrate the effectiveness of the gravity compensation strategy. Moreover, the enhanced model of the PSM arm, with adjustable center of mass locations, paves the way for more precise control and further research opportunities.

\section{CONTRIBUTION}
The project's contributions are manifold:

\textbf{1. Improved Stability and Precision:}
   - The application of gravity compensation has markedly improved the stability and precision of the MTM arm. This is crucial for performing delicate surgical tasks where even minor deviations can lead to significant errors.

\textbf{2. Enhanced Simulation Models:}
   - By incorporating gravity compensation into the Gazebo models of both the MTM and PSM, we have created a robust platform for testing and refining control strategies. These simulation models are invaluable for future research, allowing for safe and repeatable experimentation.

\textbf{3. Foundation for Future Research:}
   - The development of detailed models and control strategies provides a strong foundation for future research. These models can be used to explore new control algorithms, improve existing ones, and extend the capabilities of robotic-assisted surgical systems.

\textbf{4. Contribution to Surgical Robotics:}
   - The advancements achieved through this project contribute to the broader field of surgical robotics. By improving the control and precision of robotic systems, we enhance their utility and reliability in real-world surgical applications, ultimately improving patient outcomes.

\section{CONCLUSION}
I was successful in implementing gravity compensation on the main links of the MTM. Using the parameters derived from the least square fit, the arm was able to support itself with the correct gravity compensation. Similar gravity compensation was applied to the Gazebo model of the MTM, enabling it to be controlled as well. Finally, I created a model of the PSM in Gazebo that could be used on similar projects in the future. Overall, the project has provided a very strong basis for allowing the operator full and accurate control over the da Vinci arm.

\end{document}